\documentclass[runningheads]{llncs}
\usepackage[T1]{fontenc}
\usepackage{graphicx}
\usepackage{xcolor}
\usepackage{booktabs}
\usepackage{float}
\usepackage{amsmath, amsfonts, amssymb}
\begin{document}
\title{A Lesion-aware Edge-based Graph Neural Network for Predicting Language Ability in Patients with Post-stroke Aphasia}
\titlerunning{Lesion-aware Edge-based GNN for Post-stroke Aphasia}
%
\author{Zijian Chen\inst{1}, Maria Varkanitsa\inst{2}, Prakash Ishwar\inst{1}, Janusz Konrad\inst{1}, \\ Margrit Betke\inst{3}, Swathi Kiran\inst{2} \and Archana Venkataraman\inst{1}}
\authorrunning{Z. Chen et al.}
%
\institute{ Department of Electrical and Computer Engineering, Boston University \and Center for Brain Recovery, Boston University \and Department of Computer Science, Boston University\\ \email{\{zijianc,mvarkan,pi,jkonrad,betke,skiran,archanav\}@bu.edu}
}%
\maketitle              
\begin{abstract}
    We propose a lesion-aware graph neural network (LEGNet) to predict language ability from resting-state fMRI (rs-fMRI) connectivity in patients with post-stroke aphasia. Our model integrates three components: an edge-based learning module that encodes functional connectivity between brain regions, a lesion encoding module, and a subgraph learning module that leverages functional similarities for prediction. We use synthetic data derived from the Human Connectome Project (HCP) for hyperparameter tuning and model pretraining. We then evaluate the performance using repeated 10-fold cross-validation on an in-house neuroimaging dataset of post-stroke aphasia. Our results demonstrate that LEGNet outperforms baseline deep learning methods in predicting language ability. LEGNet also exhibits superior generalization ability when tested on a second in-house dataset that was acquired under a slightly different neuroimaging protocol. Taken together, the results of this study highlight the potential of LEGNet in effectively learning the relationships between rs-fMRI connectivity and language ability in a patient cohort with brain lesions for improved post-stroke aphasia evaluation.
    
    \keywords{Lesion-aware modeling  \and Graph neural networks \and Functional connectivity \and Data augmentation \and Aphasia prediction.}
\end{abstract}

\section{Introduction}

Stroke is one of the major causes for disability worldwide~\cite{feigin2022world}, with approximately one-third of stroke survivors affected by speech and language impairments, known as aphasia~\cite{berthier2005poststroke}. Resting-state fMRI (rs-fMRI) captures steady-state patterns of co-activation in the brain and provides a unique glimpse into the altered brain network organization due to the stroke~\cite{falconer2024resting}. Exploring this relationship is crucial for understanding the mechanisms underlying aphasia and for developing effective, personalized treatment strategies. However, developing models that can simultaneously accommodate patient-specific changes in functional connectivity due to a lesion (i.e., the stroke area) and use this information to predict generalized language impairments remains an open challenge. 

Prior studies have attempted to predict language ability using neuroimaging data. The earliest work~\cite{pustina2017enhanced} developed a stacked random forest (RF) model that performed feature selection across multiple modalities and then used these features to predict the composite Aphasia Quotient scored from the revised Western Aphasia Battery, i.e., WAB-AQ~\cite{kertesz2007western}. Another study~\cite{kristinsson2021machine} used support vector regression (SVR) to predict WAB-AQ by stacking features from functional MRI, structural MRI, and cerebral blood flow data. Recent work~\cite{chennuri2023fusion} proposed  a supervised
learning method for feature selection and fusion methods to integrate features from different modalities and predicted WAB-AQ using RF and SVR. An earlier study ~\cite{billot2022multimodal} used similar multimodal ML methods to predict treatment response, rather than baseline functionality. 
Finally, Wang et al.~\cite{wang2023topological} used persistent diagrams derived from patient rs-fMRI
to identify aphasia subtypes~\cite{wang2023topological}. 
While these studies represent seminal contributions, they largely treat the data as a ``bag of features" and do not fully capitalize on network-level information. 

We propose to address this gap with Graph neural networks (GNN), which represent the brain as a graph, where nodes correspond to regions of interest (ROIs) and edges represent functional connections between ROIs. Convolutions on the graph aggregate information from neighboring nodes or edges. They can be node-based, as seen in models like BrainGNN~\cite{li2021braingnn}, GAT~\cite{velickovic2017graph}, and GIN~\cite{xu2018powerful}, or edge-based, as formulated in the BrainNetCNN~\cite{kawahara2017brainnetcnn} and the HGCNN~\cite{huang2023heterogeneous} models. GNNs have shown superior performance compared to traditional machine learning techniques in predicting cognitive outcomes related to autism~\cite{dsouza2021m,li2021braingnn}, aging and intelligence~\cite{huang2023heterogeneous}, Alzheimer's Disease~\cite{zhang2024constructing}, and ADHD~\cite{zhao2022dynamic}. However, these applications revolve around intact brain networks, which is not the case for a large lesion caused by stroke. Previous work ~\cite{nandakumar2021automated,nandakumar2021multi} took the approach of masking out the lesioned ROIs from the input data. However, this strategy ignores the possibility of informative brain signals from around the lesion boundary. Another challenge is the limited availability of rs-fMRI data from stroke patients. One approach is to reduce the number of features in the analysis~\cite{chennuri2023fusion,kristinsson2021machine,pustina2017enhanced}. However, feature selection may inadvertently remove key information in the data, and prior studies have not been diligent about cleanly separating data used for feature selection from that used for performance evaluation~\cite{billot2022multimodal,pustina2017enhanced}. 

In this paper, we introduce a novel lesion-aware edge-based GNN model, which we call LEGNet, that uses rs-fMRI connectivity to predict language ability in patients with post-stroke aphasia. LEGNet is designed to aggregate information from neighboring edges of the brain graph, thus aligning with both the nature of rs-fMRI connectivity and the distributed interactions that contribute to language performance. We incorporate lesion information into LEGNet by encoding the stroke size and position into the model and by using this encoding to constrain the graph convolution process. To address data scarcity, we draw from the approach of~\cite{nandakumar2023deep} and develop a comprehensive data augmentation strategy that inserts an ``artificial lesion" into healthy neuroimaging data and simulates the corresponding impact on rs-fMRI connectivity and language ability. We demonstrate that LEGNet outperforms baseline deep learning methods on two in-house datasets of patients with post-stroke aphasia.  

\begin{figure}[t]
    \centering
    \includegraphics[width=0.95\textwidth]{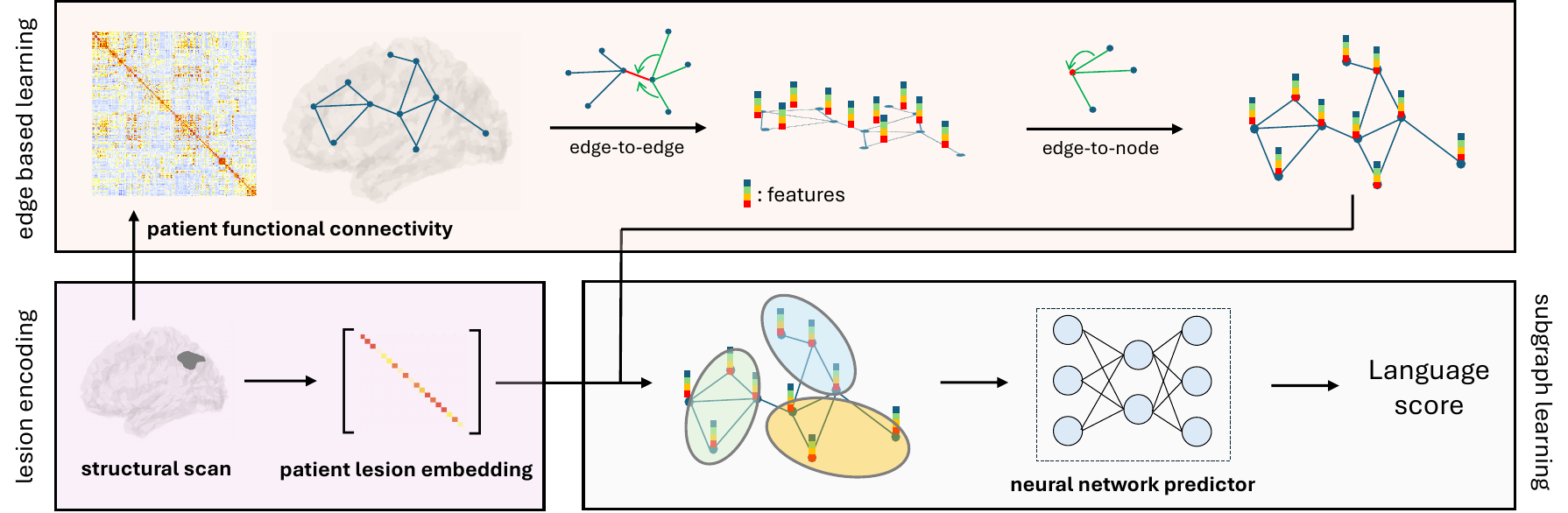}
    \caption{The Lesion-aware Edge-based BrainGNN Model. \textbf{Top:} Edge-to-edge message passing and edge-to-node aggregation. \textbf{Bottom Left:} Patient-specific lesion size and position encoding. \textbf{Bottom Right:} Subgraph updating and language prediction.}
    \label{fig:model}
\end{figure}

\section{Lesion-Aware GNN with Simulated Training Data}

\subsection{LEGNet Model Architecture}

An overview of our LEGNet model architecture is shown in Fig.~\ref{fig:model}.  LEGNet is designed to bridge the gap between the region- or node-based  characterization of a lesion and rs-fMRI connectivity, which is defined on edges. As seen, our model includes three components: an edge-based learning module, a lesion encoding module, and a subgraph learning module that connects the two viewpoints.

Formally, let $N$ be the number of ROIs in the brain. The input to LEGNet is the patient rs-fMRI connectivity~$\mathbf{X}\in \mathbb{R}^{N\times N}$, which is obtained by exponentiating the correlation matrix computed from the mean time series of non-lesioned voxels within each ROI, as introduced in~\cite{nandakumar2021automated}. If the entire ROI lies within the lesion, then the time series is zero. The entries of~$\mathbf{X}$ can be viewed as \textit{edge features} in the underlying brain graph defined on the ROIs. 

\medskip \noindent
\textbf{Edge-Based Learning:} From the input~$\mathbf{X}$, LEGNet first performs an edge-to-edge convolution \cite{kawahara2017brainnetcnn} given by the following relationship:
\begin{equation} \label{eq:e2e}
    \small
    \mathbf{H}_{ij} = \phi\bigg( \sum_{n\in \mathcal{N}(i)} \mathbf{r}_n \mathbf{X}_{in} + \sum_{n\in \mathcal{N}(j)} \mathbf{c}_n \mathbf{X}_{nj}  \bigg),
\end{equation}
where $\mathbf{H}_{ij} \in \mathbb{R}^{d_0}$ is the feature map of edge $(i,j)$, $\mathcal{N}(i)$ is the set of neighboring nodes to ROI~$i$, including~$i$ itself, $\mathbf{r}_n \in \mathbb{R}^{d_0}$ and $\mathbf{c}_n \in \mathbb{R}^{d_0}$ are the learnable filters for each node~$n$, and $\phi$ is an activation function that is applied element-wise. Intuitively, Eq.~(\ref{eq:e2e}) aggregates the connectivity information along neighboring edges that share the same end-nodes and updates the edge features accordingly.

Following this step, LEGNet maps the edge features back into the node space: 
\begin{equation} \label{eq:e2n}
    \small
    \mathbf{h}_i^{(1)} = \phi\bigg( \sum_{n\in \mathcal{N}(i)} \mathbf{g}_n \mathbf{H}_{in} + \mathbf{b}_1 \bigg), \quad i = 1,2,\ldots,N,
\end{equation}
where $\mathbf{h}_i^{(1)} \in \mathbb{R}^{d_1}$ is the feature map of node~$i$, $\mathbf{g}_n \in \mathbb{R}^{d_1\times d_0}$ is the learnable filter, and $\mathbf{b}_1\in \mathbb{R}^{d_1}$ is the learnable bias term from~\cite{kawahara2017brainnetcnn}.

\medskip \noindent
\textbf{Lesion Encoding:} The LEGNet lesion encoding module captures the size and position of the stroke for downstream processing. This is done by computing the percentage of spared gray matter~$p_i$ in each ROI~$i$. We use this information to construct the diagonal lesion embedding matrix~$\mathbf{L}\in \mathbb{R}^{N\times N}$ for each patient: 
\begin{equation} \label{eq:lesion}
    \small
    \mathbf{L} = \begin{bmatrix}
    | & | &  & | \\
    \mathbf{L}_1 & \mathbf{L}_2 & \cdots & \mathbf{L}_N \\
    | & | &  & | \\
    \end{bmatrix} = \begin{bmatrix}
        p_1 & &  \\
        & \ddots &  \\
        & & p_N
    \end{bmatrix},
\end{equation}
If an ROI is intact, then $p_i=1$ to indicate no lesion; otherwise, $0\leq p_i<1$. 

\medskip \noindent
\textbf{Subgraph Learning:} At a high level, the subgraph learning module divides the nodes/ROIs into $k$ subgroups based on their lesion encoding information and their (learned) contributions to the final prediction. First, LEGNet uses the lesion encoding~$\mathbf{L}$ to update the node representations via: 
\begin{equation} \label{eq:subgraph}
    \small
    \mathbf{h}_i^{(2)} = \phi\bigg( \sum_{j\in \mathcal{N}(i)} \mathbf{W}_j \mathbf{h}_j^{(1)} \bigg), \quad i = 1,2,\ldots,N,
\end{equation}
where $\mathbf{h}_i^{(2)}\in \mathbb{R}^{d_2}$ is the updated representation for node~$i$, and, inspired by~\cite{li2021braingnn}, the filters~$\mathbf{W}_j \in\mathbb{R}^{d_2\times d_1}$ are parameterized using the lesion matrix~$\mathbf{L}$ as follows:
\begin{equation} \label{eq:filter}
    \small
    \operatorname{vec}(\mathbf{W}_j) = \mathbf{\Theta}_2 \cdot \psi(\mathbf{\Theta}_1 \mathbf{L}_j) + \mathbf{b}_2. 
\end{equation}
The learnable parameters $\mathbf{\Theta}_2 \in \mathbb{R}^{d_2d_1 \times k}$ and $\mathbf{\Theta}_1 \in \mathbb{R}^{k \times N}$ are shared across all regions and all subjects. The bias term is $\mathbf{b}_2$, and $\psi$ an activation function.

The assignment score for each node $j$ is computed as $\psi\left(\mathbf{\Theta}_1 \mathbf{L}_j\right)$ and depends on its lesion embedding. The score indicates the involvement of node~$j$ in each subgraph. In this way, ROIs with similar lesion information and functionality are grouped together and updated with similar filters. Following the subgraph learning, the updated node features $\mathbf{h}_i^{(2)}$ are fed into a fully connected layer, with dimension~$d_3$, to predict the scalar WAB-AQ, which quantifies language ability. 

\medskip \noindent
\textbf{Training Loss:} We train LEGNet using the mean squared error between the actual~$y_m$ and predicted~$\hat{y}_m$ language performance for each subject~$m$, together with a ridge regularization term on the network filters:
\begin{equation}
    \ell = \frac{1}{M}\sum_{i=1}^M (\hat{y}_i - y_i)^2 + \lambda \,R(\mathbf{\Theta}_1, \mathbf{\Theta}_2, \mathbf{b}_1, \mathbf{b}_2, \mathbf{r}, \mathbf{c},\mathbf{g}),
\end{equation}
where $M$ is the total number of subjects and $R$ is an $L^2-$norm.

\subsection{Synthetic Data Generation for Model Pre-Training}
\label{sec:synthetic}

Given the heterogeneity of stroke, we pre-train LEGNet using a large simulated dataset. This pre-trained model is then fine-tuned using our small patient dataset. Our strategy is to insert ``artificial lesions" into the neuroimaging data of healthy subjects and simulate its impact on rs-fMRI connectivity and language. 

\begin{figure}[t]
    \centering
    \includegraphics[width=0.9\textwidth]{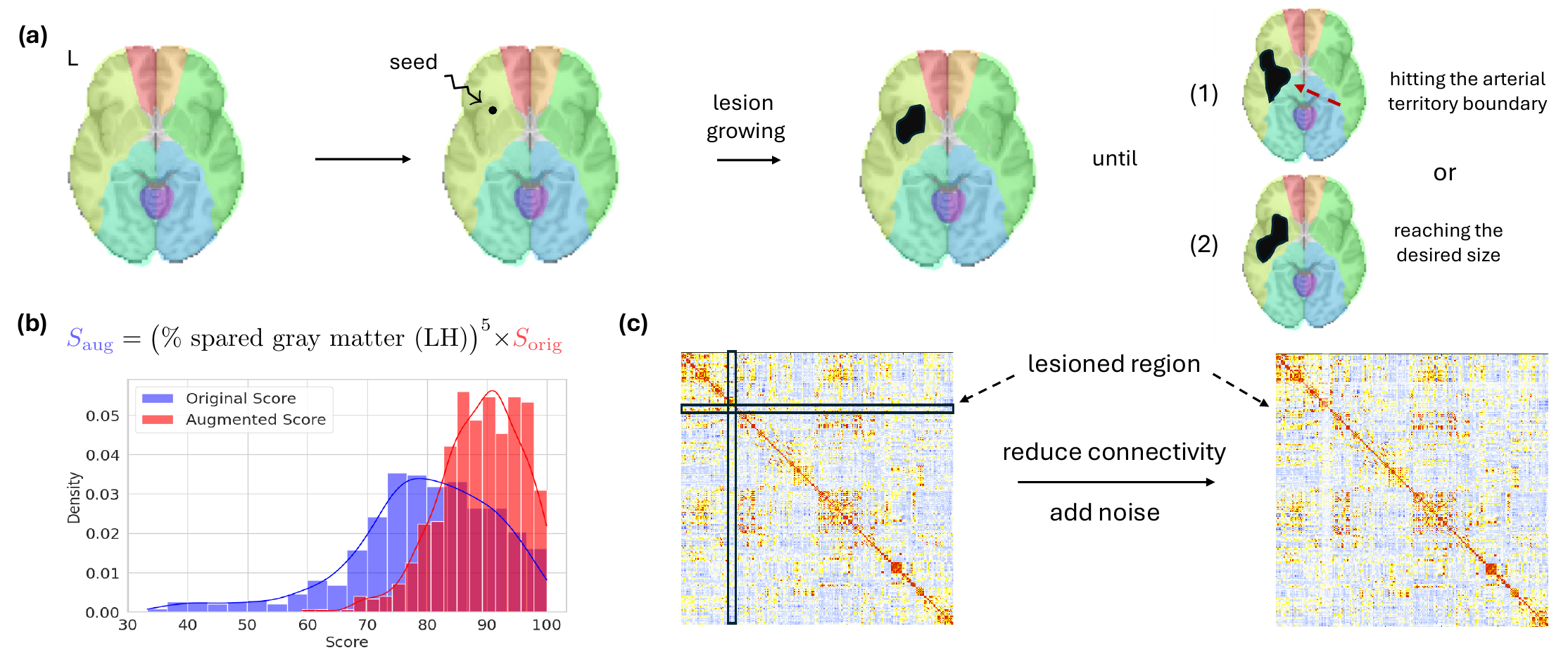}
    \caption{Synthetic data generation workflow. \textbf{(a)} Artificial lesions are created to lie within a single arterial territory. \textbf{(b)} Language score density is adjusted after the lesion augmentation. \textbf{(c)} Functional connectivity is corrupted in the lesioned area. }
    \label{fig:HCP}
\end{figure}

Our pipeline for generating synthetic data is shown in Fig.~\ref{fig:HCP}. We first simulate a unique structural lesion for each subject based on the following rules: (1) lesions are left-hemisphere only; (2) lesions are placed randomly but do not cross arterial territories \cite{liu2023digital}; (3) lesion sizes range from 5\% to 20\% of one arterial territory; (4) lesions are spatially continuous and simply-connected (i.e., without holes in the inside). Next, the artificial lesion is used to mask out voxels when computing ROI mean time series. We also diminish and add Gaussian noise to the connectivity represented in $\mathbf{X}$ between the lesioned region and the rest of the brain, followed by clipping the values to lie within the original connectivity range. Finally, the language performance score is re-scaled proportional to the percentage spared gray matter ($<1$) to simulate the negative impact of the lesion on functionality. 

\subsection{Implementation Details}

\textbf{Simulated-Lesion HCP (HCP-SL):} We use rs-fMRI data from 700 randomly selected subjects in the Human Connectome Project (HCP) S1200 database~\cite{van2013wu} as the foundation for generating synthetic data. Following the standard HCP minimal preprocessing pipeline~\cite{smith2013resting}, we parcellate the brain into 246 ROIs using the Brainnetome atlas~\cite{fan2016human}. The subject language score is accuracy in answering simple math and story-related questions during an fMRI language task. Artificial lesions are inserted and modify the data as described in Section.~\ref{sec:synthetic}.  

\medskip \noindent
\textbf{Pre-training:} 
Pre-training is done via 10-fold cross validation (CV). We use a two-stage grid search to fix the model hyperparameters~$\{\lambda, k, d_0, d_1, d_2, d_3\}$, with a coarse stage used to select a suitable power of $2$ from $2^1$ to $2^6$, followed by a fine stage with increments of 1-2. The regularizer $\lambda$ is swept across $[10^{-4},1]$. The final values are~$\{\lambda=0.005, k=8, d_0=4, d_1=8, d_2=2, d_3=8\}$.  We use the Adam optimizer with learning rate starting from 0.01 and decaying by a factor of 0.95 every 20 steps. Early stopping is also applied based on the validation loss. Once the hyperparameters are selected, we pre-train the LEGNet architecture in order to provide a better model initialization for our stroke datasets.\footnote{All code and synthetic data will be made public upon paper acceptance.}

\medskip \noindent
\textbf{Application to Post-Stroke Aphasia:} We use repeated 10-fold CV on our larger in-house dataset (see Section~3.1) to evaluate the performance of LEGNet in a real-world setting. The hyperparameters and optimizations are fixed at the values determined during the pre-training phase on synthetic data. The same procedure is applied to all baseline methods. We compare two scenarios: (1)~training the models (LEGNet and baselines) from scratch, and (2)~using the pre-trained model as the initialization for our repeated CV experiment. 

\medskip \noindent
\textbf{Cross-Dataset Generalization:} As further validation, we quantify the language ability prediction performances when the models (LEGNet and baselines) are trained on DS-1 and applied to our second in-house dataset (DS-2) which has slightly different patient characteristics than DS-1.

\subsection{Baseline Models}

We compare LEGNet with four baseline approaches. The first baseline is a modified BrainGNN model (BrainGNN$\dag$)~\cite{li2021braingnn}, which uses the same subgraph learning modules but does not perform edge-based learning or incorporate lesion information. The second baseline is BrainNetCNN model ~\cite{kawahara2017brainnetcnn} with ROIs masked from the rs-fMRI connectivity input if the percentage of spared gray matter is less than 0.3 (BNC-masked). The third baseline is the BrainNetCNN with a two-channel input, with one channel being the unaltered rs-fMRI connectivity and the second channel being a lesion mask (BNC-2channel). The final baseline is support vector regression (SVR) with the lower-triangle of the rs-fMRI connectivity matrix used as the input feature vector. 
The baseline models inherit the appropriate subset of hyperparameters from LEGnet.  We used the default radial basis function kernel with $C=100, \gamma=2/(N(N-1))$, and $\epsilon=0.1$.

\section{Experimental Results}

\subsection{Datasets of Post-Stroke Aphasia} 

\medskip \noindent
\textbf{In-House Dataset 1 (DS-1):} This dataset consists of 52 patients with chronic post-stroke aphasia with left-hemisphere lesions and aged between 35--80 years.
Structural MRI (T1-weighted; TE=2.98ms, TR=2300ms, TI=900ms, res=1mm isotropic) and rs-fMRI (EPI; TE=20ms, TR=2/2.4s, res=$1.72\times 1.72\times 3$mm$^3$) were acquired on a Siemens 3T scanner. Both scans are pre-processed using the CONN toolbox~\cite{whitfield2012conn}. Lesion boundaries were delineated manually by trained professionals and normalized to the MNI space. We use the Brainnetome atlas~\cite{fan2016human} to delineate 246 ROIs for the input rs-fMRI connectivity. Finally, all patients were evaluated using the WAB test~\cite{kertesz2007western} to obtain a measure of overall language ability (i.e., WAB-AQ). This value ranges from 0-100 with lower scores indicating severe aphasia, and higher scores indicating mild aphasia.

\medskip \noindent
\textbf{In-House Dataset 2 (DS-2):} This dataset consists of 18 patients with chronic post-stroke aphasia that were recruited separately from DS-1. While the inclusion criteria and neuroimaging acquisition protocols are the same as for DS-1, the distribution of WAB-AQ scores is different. This provides an ideal scenario to evaluate cross-dataset generalization of LEGNet and the baseline models. 

\begin{table}[t]
    \centering
    \caption{Language prediction on DS-1 using repeated 10-fold CV for LEGNet and the baselines in Section~2.4.
    The asterisk * indicates statistically worse performance ($p<0.05$) compared to the best performing model in highlighted in bold.}
    \label{tab:p50}
    \renewcommand{\arraystretch}{1.2}
    \resizebox{0.95\textwidth}{!}{%
    \begin{tabular}{ccccc}
    \toprule
    Methods        & RMSE                      &   MAE    & $R^2$                   & Correlation Coeff.     \\ 
    \midrule

    LEGNet         & \textbf{17.38 $\pm$ 0.44} &  $12.56\pm 3.02$        & \textbf{0.35 $\pm$ 0.03} & \textbf{0.61 $\pm$ 0.03} \\
    BrainGNN$\dag$       & $19.46\pm 0.30 *$         &  $14.02\pm 2.59*$        & $0.29\pm 0.04 *$          & $0.59 \pm 0.02*$      \\
    BNC-mask       & $19.18\pm 1.82 *$         &  \textbf{11.61$\pm$ 2.33}      & $0.22\pm 0.05 *$          & $0.55\pm 0.03 *$          \\
    BNC-2channel   & $22.78\pm 0.53 *$         &  $19.41\pm 0.89*$        & $0.21\pm 0.10 *$          & $0.53\pm 0.05 *$          \\
    SVR            & $20.45\pm 0.35 *$         &  $16.85\pm 0.26*$        & $0.13\pm 0.05 *$          & $0.55\pm 0.03 *$          \\ 
    \hline
    LEGNet (w/o HCP-SL) & 18.39 $\pm$ $0.68*$ &  15.33 $\pm$ $0.57*$       & 0.29 $\pm$  $0.05*$ & 0.58 $\pm$ 0.04 \\
    \bottomrule
    \end{tabular}}
\end{table}

\subsection{Performance Characterization and Model Interpretation}

Table~\ref{tab:p50} reports the predictive performance of each method using repeated 10-fold CV on DS-1. LEGNet achieves the best performance in RMSE, $R^2$, and correlation coefficient. While it is second-best to BNC-masked in MAE, the difference is not statistically significant. As a baseline, we applied the LEGNet architecture to DS-1 from a random initialization, i.e., without having access to synthetic data. To avoid data leakage, we selected the hyperparameters based on the corresponding modules used in previous studies~\cite{li2021braingnn,nandakumar2021automated}. We note a statistically significant decrease in performance w/o HCP, which underscores the importance of using synthetic data to design and initialize the deep network. 

\begin{figure}[t]
    \centering
    \includegraphics[width=0.95\textwidth]{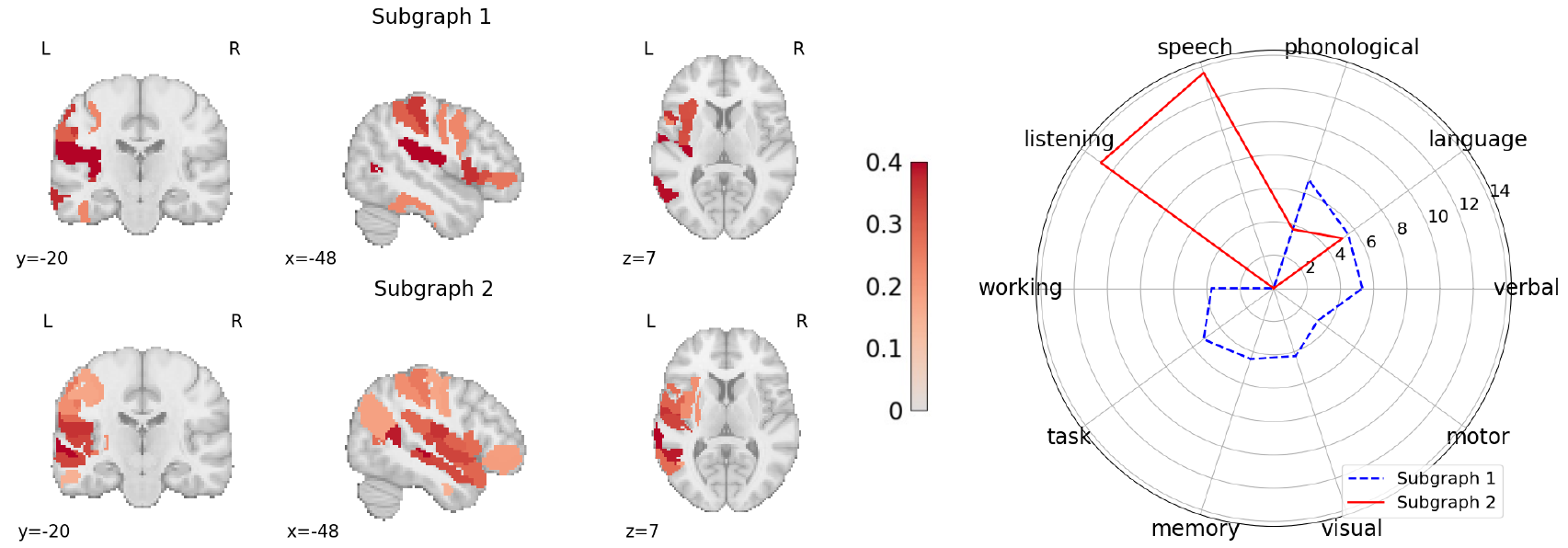}
    \caption{\textbf{Left:} Top two subgraphs in in DS-1. \textbf{Right:} The $z-$score of each subgraph from Neurosynth~\cite{yarkoni2011large} indicating the  association strength with a particular function.}
    \label{fig:assignment}
\end{figure}

Fig.~\ref{fig:assignment} (left) illustrates the top two subgraphs identified by LEGNet for the best-performing model during repeated 10-fold CV. The top subgraphs are identified by averaging the subgraph assignment scores for each ROI ($\psi\left(\mathbf{\Theta}_1 \mathbf{L}_j\right)$ from Section~2.1) across all 52 patients in DS-1. We use Neurosynth~\cite{yarkoni2011large} to decode the functionality associated with the ROIs assigned to each of the top two subgraphs, as shown in Fig.~\ref{fig:assignment} (right). We note that LEGNet assigns high scores to regions that are related to the language ability. Intuitively, these regions also influence the prediction of language ability, as described in Section~2.1.

\begin{table}[t]
    \setlength{\tabcolsep}{7pt}
    \centering
    \caption{Language prediction on DS-2 with the model from repeated 10-fold CV that generalizes the best. Top performance is highlighted in bold.}
    \label{tab:md}
    \begin{tabular}{ccccc}
    \toprule
    Methods        & RMSE                      &   MAE    & $R^2$                   & Correlation Coeff.     \\ 
    \midrule
    
    LEGNet                  & \textbf{17.71}    & \textbf{8.74}    & \textbf{0.19}  & \textbf{0.44} \\
    BrainGNN$\dag$               & 18.52             & 12.64            & 0.11           & 0.34 \\
    BNC-mask                & 19.24             & 12.68            & 0.04           & 0.31 \\
    BNC-2channel            & 19.70             & 13.47            & 0.01           & 0.28 \\
    SVR                     & 18.43             & 12.65            & 0.12           & 0.35 \\
    \hline
    LEGNet (w/o HCP-SL)     & 18.36             & 11.26            & 0.12           & 0.40 \\
    \bottomrule\\
    \end{tabular}
\end{table}

Finally, we tested generalization performance by applying the model that performs best on DS-1 to DS-2 without any fine-tuning
(Table~\ref{tab:md}). In terms of $R^2$, LEGNet maintains a leading position but, along with the other baselines, also shows a decrease compared to the validation performance in Table~\ref{tab:p50}. While BrainGNN and SVR also show a decrease, BrainNetCNN-based models exhibit a sharper drop, indicating their reduced robustness 
on unseen data. The other three metrics follow a similar trend. This is expected due to the slight distribution shift between DS-1 and DS-2. Nevertheless, LEGNet still outperforms all baseline methods, indicating superior generalization ability.

\section{Conclusion}
We have introduced LEGNet, a novel lesion-aware edge-based graph neural network model designed to predict language performance in post-stroke aphasia patients from rs-fMRI connectivity. LEGNet bridges the gap between the lesion boundary defined on nodes and rs-fMRI connectivity defined on edges, while simultaneously using the lesion size and position to guide both the graph convolution and subgraph identification processes. Our synthetic data generation procedure addresses the challenge of limited patient data by simulating lesioned brain networks in healthy subjects. Pretraining on the augmented HCP dataset allows for unbiased hyperparameter selection and a reliable model initialization for fine-tuning on patient data. We demonstrate that LEGNet outperforms state-of-the-art methods in predictive accuracy and generalization ability, thus highlighting its potential as a reliable tool for post-stroke aphasia evaluation. 

\subsection*{Acknowledgements}
This work was supported by the National Institutes of Health R01 HD108790 (PI Venkataraman), the National Institutes of Health R01 EB029977 (PI Caffo), the National Institutes of Health R21 CA263804 (PI Venkataraman), the National Institutes of Health P50DC012283 (BU Site PI Kiran) and the National Institutes of Health R01 DC016950 (PI Kiran).

%
%
%
\bibliographystyle{splncs04}
\bibliography{refs}

\end{document}